\documentclass[runningheads]{llncs}
\usepackage{graphicx}
\usepackage{amsmath,amssymb}

\usepackage{color}

\begin{document}
\title{Multi-view Cross-Modality MR Image Translation for Vestibular Schwannoma and Cochlea Segmentation}
\titlerunning{Multi-view Cross-Modality MR Image Translation}
\author{
Bogyeong Kang \and
Hyeonyeong Nam \and
Ji-Wung Han \and
\\Keun-Soo Heo \and
Tae-Eui Kam\thanks{Corresponding author.}
}
\authorrunning{B. Kang et al.}
\institute{
Department of Artificial Intelligence, Korea University, Seoul, Republic of Korea \\ \email{\{kangbk,kamte\}@korea.ac.kr}}
\maketitle              % typeset the header of the contribution
\begin{abstract}
In this work, we propose a multi-view image translation framework, which can translate contrast-enhanced $\text{T}_1$ (ce$\text{T}_1$) MR imaging to high-resolution $\text{T}_2$ (hr$\text{T}_2$) MR imaging for unsupervised vestibular schwannoma and cochlea segmentation. We adopt two image translation models in parallel that use a pixel-level consistent constraint and a patch-level contrastive constraint, respectively. Thereby, we can augment pseudo-hr$\text{T}_2$ images reflecting different perspectives, which eventually lead to a high-performing segmentation model. Our experimental results on the CrossMoDA challenge show that the proposed method achieved enhanced performance on the vestibular schwannoma and cochlea segmentation.


\end{abstract}

\section{Introduction}
Vestibular schwannoma (VS) is a benign tumor that occurs in the nerve membrane cells of the vestibular nerve~\cite{dorent2022crossmoda,shapey2021segmentation}. For diagnosis and treatment of VS, it is necessary to segment the VS and its surrounding organs, especially the cochleas~\cite{dorent2022crossmoda,shapey2021segmentation}. In general, VS is diagnosed through contrast-enhanced $\text{T}_1$ (ce$\text{T}_1$) MR imaging but there are concerns about side effects such as allergy to gadolinium-containing contrast agents~\cite{dorent2022crossmoda,shapey2021segmentation}. As an alternative, high-resolution $\text{T}_2$ (hr$\text{T}_2$) MR imaging, a non-contrast imaging technique, has shed light on VS segmentation~\cite{dorent2022crossmoda,shapey2021segmentation}. However, it is very time-consuming and expensive to manually annotate newly released data. For this reason, the lack of annotated data can be a big problem for applying deep learning techniques in the medical domain. This issue can be solved by applying unsupervised domain adaptation, which allows a model trained in one domain to be adapted in another unseen domain without supervision~\cite{chen2019synergistic,dorent2020scribble,huo2018synseg}. Recently, some studies~\cite{shin2022cosmos,dong2021unsupervised,choi2021using} have been conducted based on cross-modality domain adaptation for VS and cochlea segmentation in unseen hr$\text{T}_2$ scans. Previous studies~\cite{shin2022cosmos,dong2021unsupervised,choi2021using} achieved outstanding performance on VS and cochlea segmentation utilizing image translation models such as CycleGAN~\cite{zhu2017unpaired} or CUT~\cite{park2020contrastive}. Of note, CycleGAN employs pixel-level consistent constraints, while CUT adopts patch-level contrastive constraints. The former constraint can better reflect the intensity and the texture of VS through cycle-consistency loss, but the structure of VS and cochleas could be distorted. Besides, the latter constraint uses contrastive loss, having an advantage in preserving the structure of VS and cochleas, but could ignore the detailed characteristics such as intensity and texture. Based on these considerations, we believe that we can obtain diverse pseudo-hrT2 images, which can help to improve the segmentation model performance by using the two aforementioned constraint models together.

Therefore, we design a multi-view image translation framework to obtain the pseudo-hr$\text{T}_2$ images with different perspectives by adopting two image translation models in parallel, CycleGAN~\cite{zhu2017unpaired} and QS-Attn~\cite{hu2022qs}. CycleGAN employs a pixel-level consistent constraint, and QS-Attn is an advanced patch-level contrastive constraint method that focuses on domain-relevant features~\cite{hu2022qs}. To our best knowledge, QS-Attn~\cite{hu2022qs} is first adopted for image translation from ce$\text{T}_1$ to hr$\text{T}_2$ images in this work. Based on our multi-view image translation framework, the following segmentation model can learn both structure and texture of VS and cochleas.

\section{Related Work}

Cross-modality unsupervised domain adaptation has drawn a lot of attention in the CrossMoDA challenge~\cite{dorent2022crossmoda}. The goal of this challenge is to construct a VS and cochlea segmentation model on hr$\text{T}_2$ images with unpaired annotated ce$\text{T}_1$ and non-annotated hr$\text{T}_2$ scans. Recent studies~\cite{shin2022cosmos,dong2021unsupervised,choi2021using} first translated the source ce$\text{T}_1$ images to the target hr$\text{T}_2$ images, and then trained their segmentation models with the translated hr$\text{T}_2$ (i.e., pseudo-hr$\text{T}_2$) images. More specifically, Shin et al.~\cite{shin2022cosmos} translated the ce$\text{T}_1$ images to the hr$\text{T}_2$ images by adding an additional decoder to CycleGAN to preserve the structures of  VS and cochleas. Dong et al.~\cite{dong2021unsupervised} conducted image translation using NiceGAN~\cite{chen2020reusing}, which is based on CycleGAN~\cite{zhu2017unpaired}, and Choi et al.~\cite{choi2021using} obtained pseudo-hr$\text{T}_2$ images using CUT~\cite{park2020contrastive}. Of note, they all obtained pseudo-hr$\text{T}_2$ images by taking only one constraint model. Besides, Choi et al.~\cite{choi2021using} performed post-processing to obtain the images with low intensity, similar to the VS in real hr$\text{T}_2$ scans.

\section{Proposed Method}
\subsection{Overview}
\begin{figure}[t]
	\includegraphics[width=\textwidth]{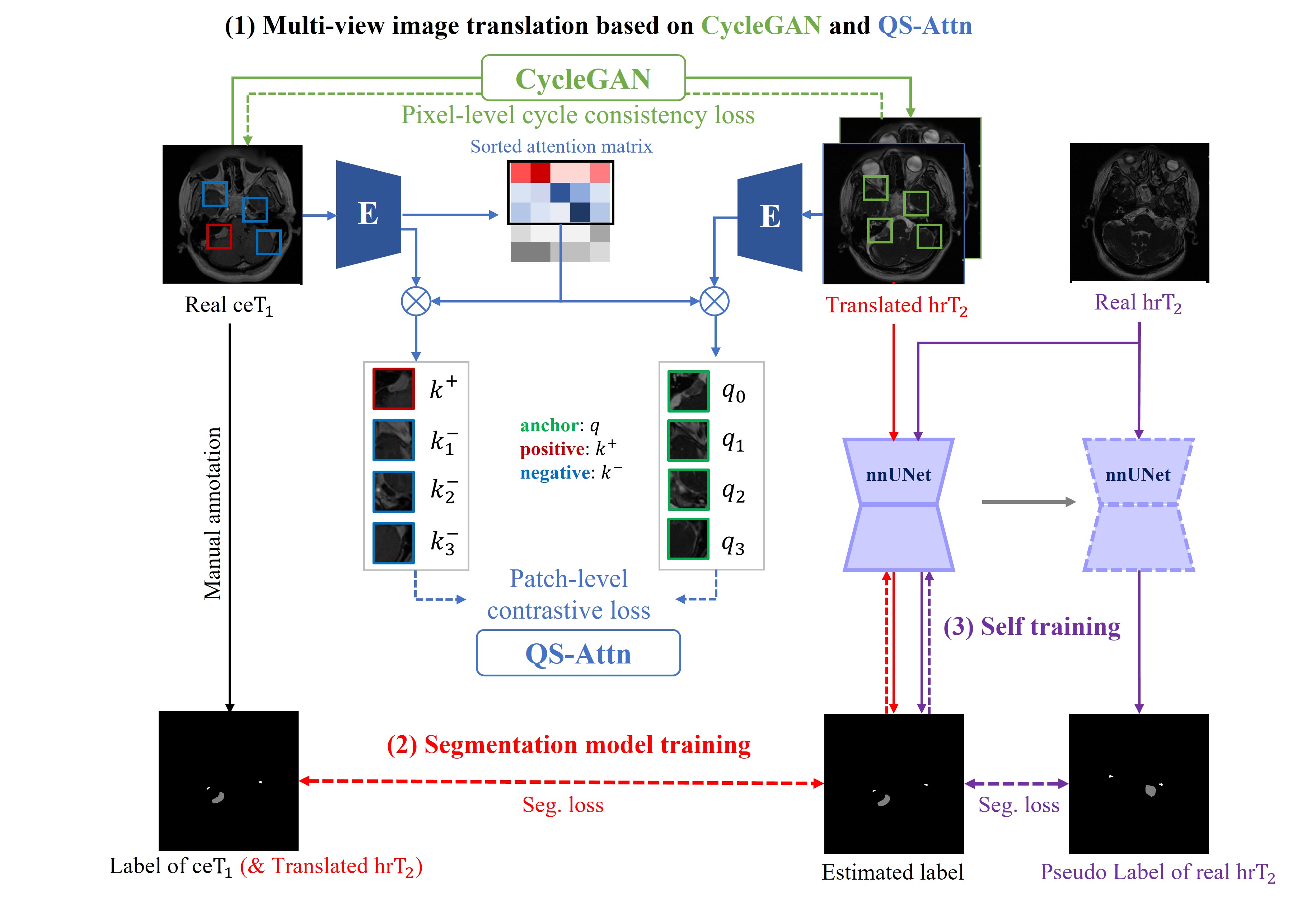}
	\caption{The overview of our proposed framework.}
	\label{fig1}
\end{figure}

Fig.~\ref{fig1} shows an overview of our proposed framework, which consists of three parts; (1) multi-view image translation, (2) segmentation model training, and (3) self-training. Specifically, we first generate the pseudo-hr$\text{T}_2$ images with various characteristics through multi-view image translation. After that, we train the segmentation model using the multi-view pseudo-hr$\text{T}_2$ images and the labels of the ce$\text{T}_1$ images. In the self-training, the trained segmentation model first performs pseudo-labeling of real hr$\text{T}_2$  images, and then is further trained by including the pseudo-labeled real hr$\text{T}_2$  images in the next training phase.

\subsection{Multi-view image translation}
We first translate ce$\text{T}_1$ images into multi-view pseudo-hr$\text{T}_2$ images by adopting CycleGAN~\cite{zhu2017unpaired} and QS-Attn~\cite{hu2022qs} in parallel.

\subsubsection{CycleGAN.} CycleGAN uses cycle-consistency loss to translate the source domain ce$\text{T}_1$ images into the target domain hr$\text{T}_2$ images. Cycle-consistency loss described in Eq.~\ref{eq7} encourages $F(G({x}_s ))$ to be equal to ${x}_s$ and $G(F({x}_t ))$ to be equal to ${x}_t$ in pixel-level when given the $G:{X}_s\to{X}_t$  and $F:{X}_t\to{X}_s$ generators~\cite{zhu2017unpaired}.
\begin{equation}
    {L}_{cycle}=\left\|F(G({x}_s)) - {x}_s\right\| + \left\|G(F({x}_t )) - {x}_t\right\|\label{eq7}
\end{equation}

\subsubsection{QS-Attn.} QS-Attn is an unpaired image translation model that is improved from CUT~\cite{park2020contrastive}. CUT preserves the structural information by constraining the patches from the same location on the source and the translated images to be close, compared to the different locations. CUT maximizes the mutual information between the source and translated images through the Eq.~\ref{eq8} ~\cite{park2020contrastive}, 
\begin{equation}
    {L}_{con}=-\log\Biggl[\cfrac{\exp(q\cdot{k}^{+}/\tau)}{\exp(q\cdot{k}^{+}/\tau)+\sum\nolimits_{i=1}^{N-1}\exp(q\cdot{k}^{-}/\tau)}\Biggr]\label{eq8}
\end{equation}

\noindent{where $q$ is the anchor feature from the translated image and ${k}^+$ is a single positive at the same location in the source image and ${k}^-$ are $(N-1)$ negatives at the other locations, and $\tau$ is a temperature~\cite{hu2022qs}.}

However, CUT~\cite{park2020contrastive} calculates the contrastive loss between the randomly selected  patches, which could have less domain-relevant information. QS-Attn addresses this limitation by adopting the QS-Attn module, which can select domain-relevant patches. The QS-Attn module constructs the attention matrix ${A}_g$ using the features in the source images and then obtains the entropy ${H}_g$ by following Eq.~\ref{eq9} ~\cite{hu2022qs}.

\begin{equation}
    {H}_g(i)=-\sum_{j=1}^{HW}{A}_g(i,j)\log{{A}_g(i,j)}\label{eq9}
\end{equation}

\noindent{Of note, the smaller entropy ${H}_g$ means the more important feature. Thus, ${A}_g$ is sorted in ascending order according to entropy ${H}_g$ to select domain-relevant patches~\cite{hu2022qs}. By calculating the contrastive loss using the selected domain-relevant patches, the structures of the source domain better preserve, and more realistic images are generated compared to CUT~\cite{park2020contrastive}.}

We empirically found that CycleGAN with pixel-level cycle-consistency loss allows the model to better reflect the intensity and the texture of the VS and cochleas in the target images, while QS-Attn takes advantage of preserving the structure of them more clearly via patch-level contrastive loss (refer to Section~\ref{discussion} for more details). By using them together, our multi-view image translation can augment pseudo-hr$\text{T}_2$ images from different perspectives, and it can help improve the performance of the following segmentation model.

\subsection{Segmentation and Self-training}
Motivated by the previous works~\cite{shin2022cosmos,dong2021unsupervised,choi2021using}, we also utilize nnUNet~\cite{isensee2021nnu} and self-training procedure~\cite{xie2020self} to construct the segmentation model. nnUNet is a powerful segmentation framework that automatically performs pre-processing, training, and post-processing with heuristic rules~\cite{isensee2021nnu}. Self-training is carried out to reduce the distribution gap between real hr$\text{T}_2$ and translated hr$\text{T}_2$ images and to improve the robustness of the segmentation model for unseen real hr$\text{T}_2$ scans. The segmentation and self-training procedure consists of four steps; (1) training the segmentation model using the translated hr$\text{T}_2$ scans with labels of the ce$\text{T}_1$ scans. (2) Generating pseudo labels of unlabeled real hr$\text{T}_2$ scans by using the trained segmentation model. (3) Retraining the segmentation model using both the translated hr$\text{T}_2$ scans with labels of the ce$\text{T}_1$ scans and the real hr$\text{T}_2$ scans with pseudo labels. 4) Repeating Steps 2-3 to achieve further performance improvement.

\section{Experiments and Results}

\subsection{Dataset and preprocessing}
We used the CrossMoDA dataset~\footnote{https://crossmoda-challenge.ml/}~\cite{dorent2022crossmoda} for training, validation. The CrossMoDA dataset consists of data from two different institutions: London and Tilburg. The London data consists of 105 ce$\text{T}_1$ scans and 105 hr$\text{T}_2$ scans. The ce$\text{T}_1$ scans were acquired with the in-plane resolution of 0.4×0.4mm, in-plane matrix of 512×512, and slice thickness of 1.0 to 1.5 mm with an MPRAGE sequence (TR=1900 ms, TE=2.97 ms, TI=1100 ms). Meanwhile, hr$\text{T}_2$ scans were acquired with the in-plane resolution of 0.5×0.5mm, in-plane matrix of 384×384 or 448×448, and slice thickness of 1.0 to 1.5 mm with a 3D CISS or FIESTA sequence (TR=9.4 ms, TE=4.23ms). For the Tilburg data set, ce$\text{T}_1$ scans and hr$\text{T}_2$ scans consist of 105 subjects each. The ce$\text{T}_1$ scans were acquired with the in-plane resolution of 0.8×0.8mm, in-plane matrix of 256×256, and slice thickness of 1.5 mm with a 3D-FFE sequence (TR=25 ms, TE=1.82 ms). The hr$\text{T}_2$ scans were acquired with the in-plane resolution of 0.4×0.4mm, in-plane matrix of 512×512, and slice thickness of 1.0 mm with a 3D-TSE sequence (TR=2700 ms, TE=160 ms, ETL=50)~\cite{dorent2022crossmoda}. The training dataset of the $\text{CrossMoDA2022 Challenge  }^1$ contains a total of 210 ce$\text{T}_1$ scans with annotation labels and 210 hr$\text{T}_2$ scans without annotation labels. In addition, they provide 64 scans of hr$\text{T}_2$ images for validation.

Since the voxel spaces vary across scans, all the images were resampled to [0.41, 0.41, 1.5] voxel sizes. For image translation, the 3D MRI images were sliced into a series of 2D images along the axial plane and the images were center-cropped and resized to 256 × 256. After performing image translation, the translated hr$\text{T}_2$ images were merged into 3D MR imaging and fed into the segmentation model.

\subsection{Implementation details} 
We implement CycleGAN~\cite{zhu2017unpaired}, QS-Attn~\cite{hu2022qs}, and nnUNet~\cite{isensee2021nnu}, following their default parameter settings. We also apply a global attention in QS-Attn~\cite{hu2022qs}, and ensemble selection in nnUNet~\cite{isensee2021nnu} for the final prediction. All the implementations are powered by RTX 3090 24GB GPUs. The training of CycleGAN, QS-Attn, and nnUNet is performed with PyTorch 1.8.0, 1.7.1, and 1.10.2, respectively.

\subsection{Results}
\begin{figure}[hbt!]
	\includegraphics[width=\textwidth]{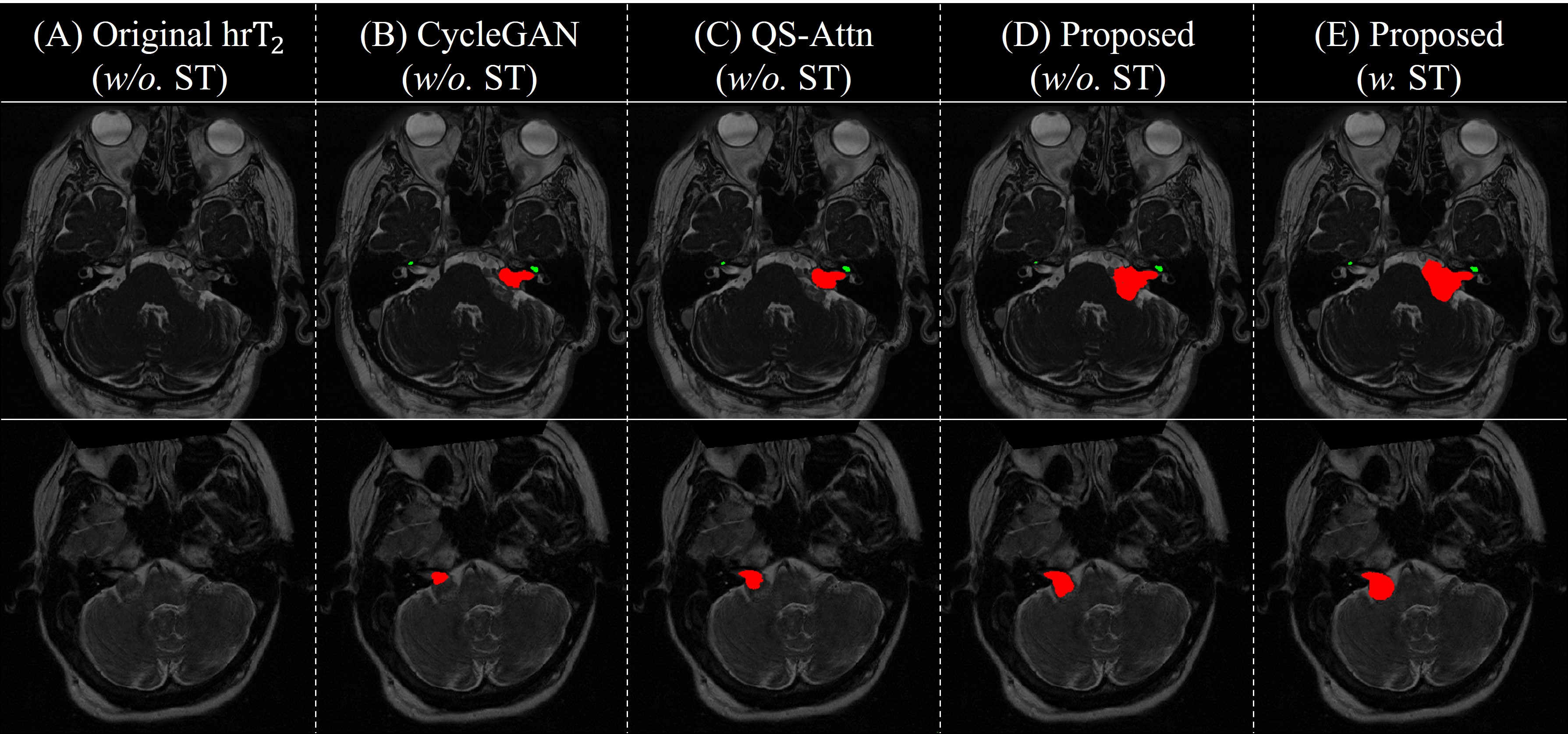}
	\caption{Qualitative comparison of segmentation results for validation set. We visualize the segmentation results of VS (red) and cochlea (green) (ST: Self-training).}
	\label{fig2}
\end{figure}

Table~\ref{tab5} and Fig.~\ref{fig2} show the VS and cochlea segmentation results with different image translation methods. The proposed multi-view image translation framework with CycleGAN~\cite{zhu2017unpaired} and QS-Attn~\cite{hu2022qs} shows better performance compared to other methods using each model alone. Moreover, we greatly improved the performance of the segmentation model with self-training. As a result, our proposed method obtained a great achievement with a mean dice score of 0.8504±0.0466 in the validation period.

\begin{table}[t]
\centering
\caption{Segmentation results with dice and ASSD scores (ST: self-training).}
\label{tab5}
\resizebox{\textwidth}{!}{%
\begin{tabular}{c|ccc|cc}
\hline
Translation                                             & \multicolumn{3}{c|}{Dice score}                                                                                                                                                                                                   & \multicolumn{2}{c}{ASSD}                                                                                                                   \\ \cline{2-6} 
model                                                   & \multicolumn{1}{c|}{VS}                                                        & \multicolumn{1}{c|}{Cochlea}                                                         & Mean                                                         & \multicolumn{1}{c|}{VS}                                                         & Cochlea                                                         \\ \hline
\begin{tabular}[c]{@{}c@{}}CycleGAN\\ (\emph{w/o.} ST) \end{tabular}                      & \multicolumn{1}{c|}{\begin{tabular}[c]{@{}c@{}}0.7798\\ ($\pm{0.1901}$)\end{tabular}} & \multicolumn{1}{c|}{\begin{tabular}[c]{@{}c@{}}0.8066\\ ($\pm{0.0323}$)\end{tabular}} & \begin{tabular}[c]{@{}c@{}}0.7932\\ ($\pm{0.0972}$)\end{tabular} & \multicolumn{1}{c|}{\begin{tabular}[c]{@{}c@{}}0.8750\\ ($\pm{0.9222}$)\end{tabular}} & \begin{tabular}[c]{@{}c@{}}0.2422\\ ($\pm{0.1608}$)\end{tabular} \\ \hline
\begin{tabular}[c]{@{}c@{}}QS-Attn\\ (\emph{w/o.} ST) \end{tabular}                       & \multicolumn{1}{c|}{\begin{tabular}[c]{@{}c@{}}0.7779\\ ($\pm{0.1825}$)\end{tabular}} & \multicolumn{1}{c|}{\begin{tabular}[c]{@{}c@{}}0.8158\\ ($\pm{0.0287}$)\end{tabular}} & \begin{tabular}[c]{@{}c@{}}0.7968\\ ($\pm{0.0929}$)\end{tabular} & \multicolumn{1}{c|}{\begin{tabular}[c]{@{}c@{}}0.6667\\ ($\pm{0.3891}$)\end{tabular}} & \begin{tabular}[c]{@{}c@{}}0.2365\\ ($\pm{0.1573}$)\end{tabular} \\ \hline
\begin{tabular}[c]{@{}c@{}}Proposed\\ (\emph{w/o.} ST) \end{tabular}                      & \multicolumn{1}{c|}{\begin{tabular}[c]{@{}c@{}}0.8043\\ ($\pm{0.1656}$)\end{tabular}} & \multicolumn{1}{c|}{\begin{tabular}[c]{@{}c@{}}0.8158\\ ($\pm{0.0289}$)\end{tabular}} & \begin{tabular}[c]{@{}c@{}}0.8101\\ ($\pm{0.0863}$)\end{tabular} & \multicolumn{1}{c|}{\begin{tabular}[c]{@{}c@{}}0.5742\\ ($\pm{0.2461}$)\end{tabular}} & \begin{tabular}[c]{@{}c@{}}0.2387\\ ($\pm{0.1581}$)\end{tabular} \\ \hline
\begin{tabular}[c]{@{}c@{}}Proposed\\ (\emph{w.} ST) \end{tabular}                      & \multicolumn{1}{c|}{\begin{tabular}[c]{@{}c@{}}\textbf{0.8520}\\ \textbf{($\pm{0.0889}$)}\end{tabular}} & \multicolumn{1}{c|}{\begin{tabular}[c]{@{}c@{}}\textbf{0.8488}\\ \textbf{($\pm{0.0235}$)}\end{tabular}} & \begin{tabular}[c]{@{}c@{}}\textbf{0.8504}\\ \textbf{($\pm{0.0466}$)} \end{tabular} & \multicolumn{1}{c|}{\begin{tabular}[c]{@{}c@{}}\textbf{0.4748}\\ \textbf{($\pm{0.2072}$)} \end{tabular}} & \begin{tabular}[c]{@{}c@{}}\textbf{0.1992}\\ \textbf{($\pm{0.1524}$)}\end{tabular} \\ \hline
\end{tabular}%
}
\end{table}

\begin{figure}[hbt!]
	\includegraphics[width=\textwidth]{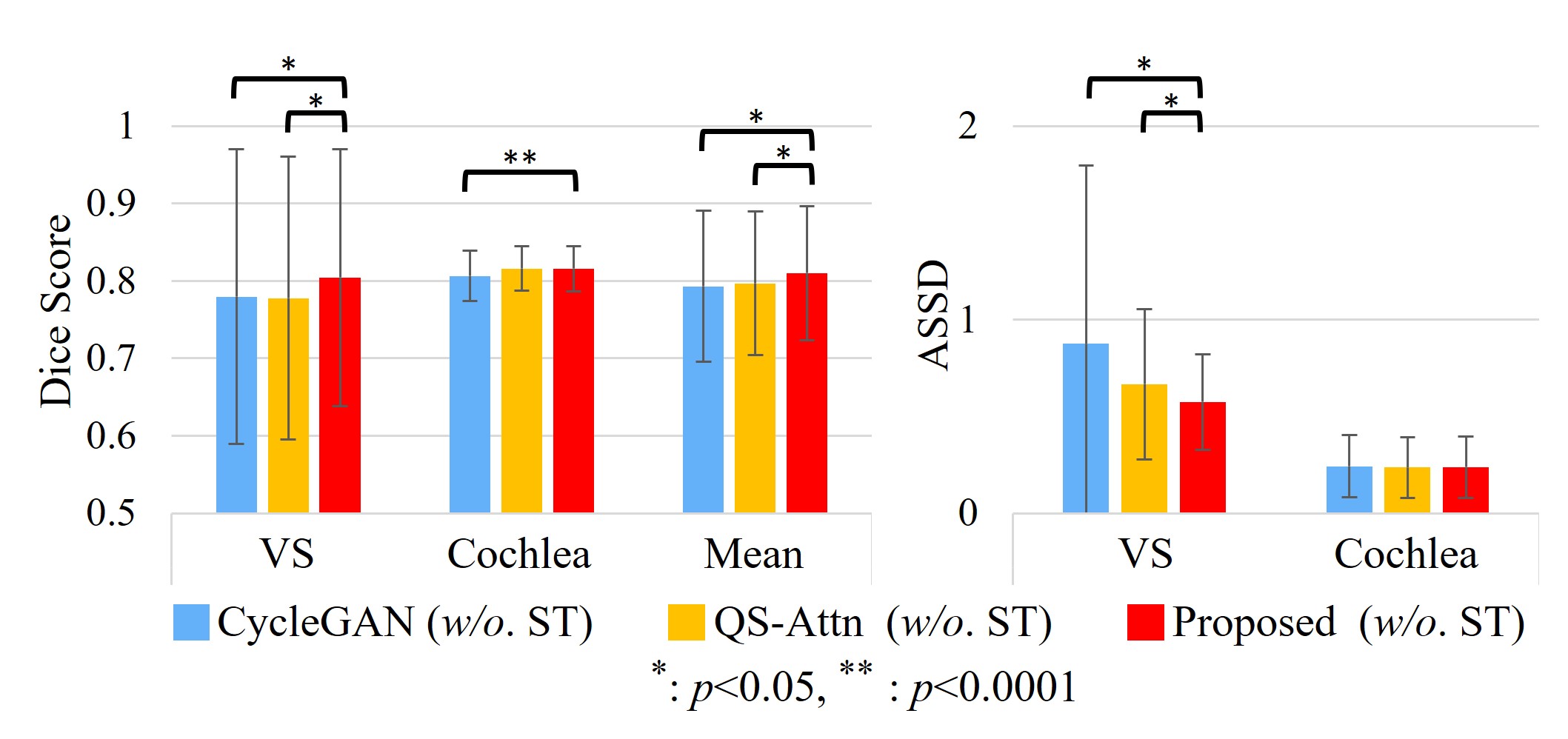}
	\caption{Performance comparison of VS and cochlea segmentation models (ST: Self-training).}
	\label{fig3}
\end{figure}

We conducted paired \emph{t}-test among CycleGAN~\cite{zhu2017unpaired}, QS-Attn~\cite{hu2022qs}, and our proposed method (\emph{w/o.} self-training, ST) to compare the segmentation performance, and the results are plotted in Fig.~\ref{fig3}. CycleGAN, QS-Attn, and our proposed method (\emph{w/o.} ST) show statistical significance with \emph{p} $<$ 0.05 for the dice score of VS and mean values. In addition, our proposed method (\emph{w/o.} ST) is statistically better with \emph{p} $<$ 0.0001 than CycleGAN on the dice score of cochleas. Through this statistical comparison, we proved that our proposed framework achieved better performance compared to other methods that use either of the two models alone.

\newpage
\section{Discussion}\label{discussion}
\begin{figure}[t]
	\includegraphics[width=\textwidth]{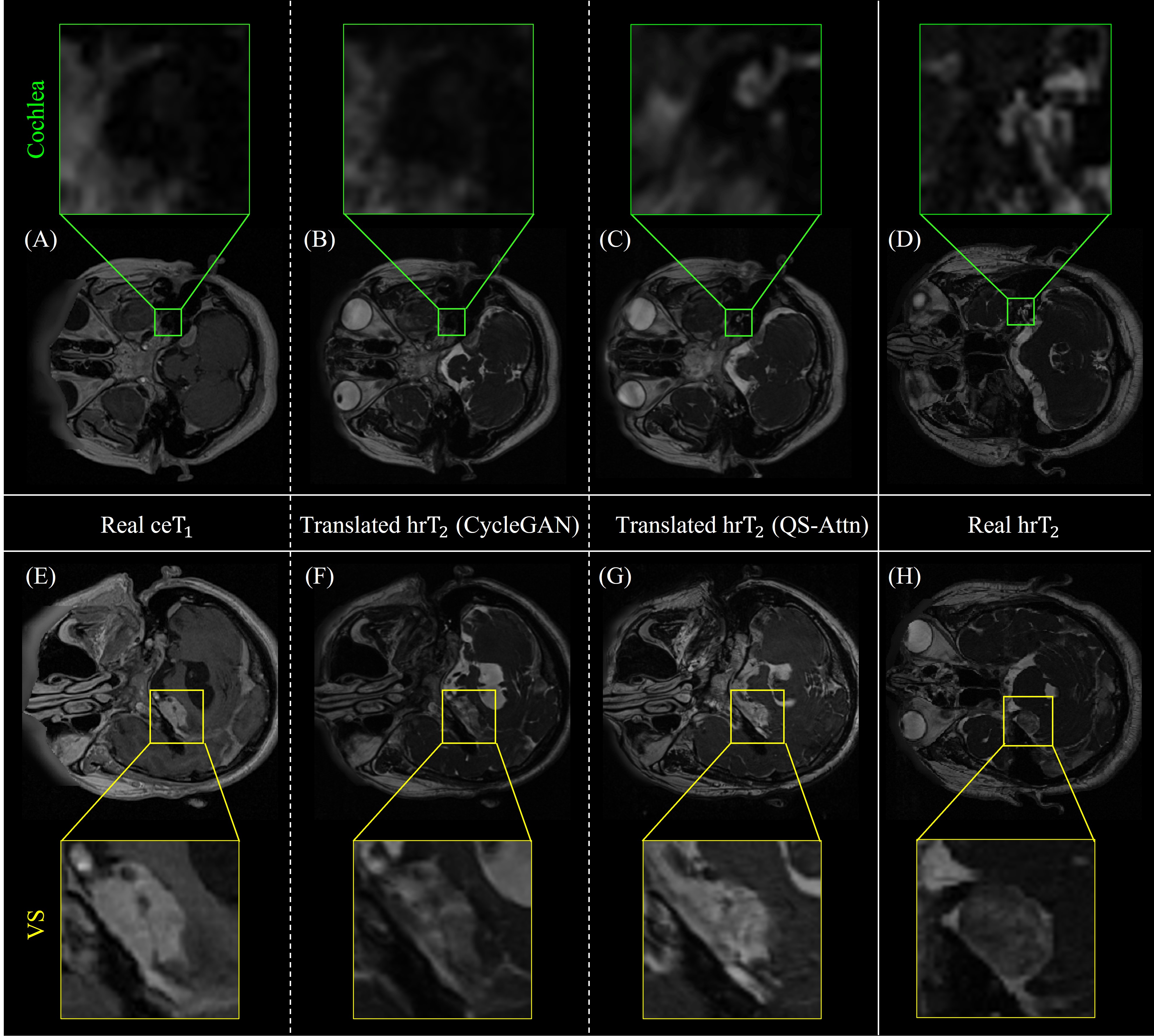}
	\caption{Comparison results of image translation by CycleGAN and QS-Attn.}
	\label{fig4}
\end{figure}

Fig.~\ref{fig4} shows the results of the two separate image translation models utilized in the multi-view image translation framework. For comparison, we randomly picked two ce$\text{T}_1$ images (A\&E), their corresponding translated hr$\text{T}_2$ images (B-C\&F-G), and two unpaired real hr$\text{T}_2$ images (D\&H). We can see that QS-Attn (C) well captured the structure of cochleas with less distortion or blurring compared to CycleGAN (B). Meanwhile, some images translated through QS-Attn (G) have too high intensities for VS, whereas those by CycleGAN (F) have similar intensity and textures to VS in the real hr$\text{T}_2$  image (H). As shown in Fig.~\ref{fig4}, the two constraint models have different strengths. Therefore, in the proposed method, the segmentation model can learn both structures and textures of VS and cochleas through our multi-view image translation framework. It allows the segmentation model to consider various perspectives of VS and cochleas and helps improve the performance of the segmentation model.

\section{Conclusion}
In this work, we design a multi-view image translation framework for VS and cochlea segmentation. Specifically, we adopt CycleGAN and QS-Attn in parallel to translate the given ceT1 images to pseudo-hrT2 images reflecting various perspectives. Based on the pseudo-hrT2 images, the segmentation model can learn both structures and textures of VS and cochleas. Our proposed method obtained great achievement in the CrossMoDA challenge2022.

\newpage
\subsubsection{Acknowledgment.}
This work was supported by Institute of Information \& communications Technology Planning \& Evaluation (IITP) grant funded by the Korea government (MSIT) (No. 2019-0-00079, Artificial Intelligence Graduate School Program (Korea University)), and the National Research Foundation of Korea (NRF) grant funded by the Korea government (MSIT) (No. 2020R1C1C1013830, No. 2020R1A4A1018309).

\bibliographystyle{splncs04}
\bibliography{ref}

\end{document}